\documentclass[sigconf,9pt]{acmart}
\renewcommand\footnotetextcopyrightpermission[1]{} 
\usepackage{caption,subcaption}
\usepackage{cleveref}

\crefformat{section}{\S#2#1#3} 
\crefformat{subsection}{\S#2#1#3}
\crefformat{subsubsection}{\S#2#1#3}

\usepackage{amsmath}
\usepackage{algorithm}
\usepackage{algpseudocode}

\usepackage{pifont}

\usepackage{multirow}            
\usepackage[table]{xcolor}       
\usepackage{siunitx}             
\author{Cheng Jiang, Yihe Yan, Yanxiang Wang, Chun Tung Chou, Wen Hu}
\affiliation{
    \institution{
    School of Computer Science and Engineering, 
    University of New South Wales}
    \city{Sydney}
    \country{Australia}
}
\email{{cheng.jiang1, yihe.yan, yanxiang.wang, c.t.chou, wen.hu}@unsw.edu.au}

\AtBeginDocument{%
  \providecommand\BibTeX{{%
    \normalfont B\kern-0.5em{\scshape i\kern-0.25em b}\kern-0.8em\TeX}}}

\definecolor{re}{RGB}{0,0,0}
\definecolor{rectc}{RGB}{0,0,0}

\begin{document}

\title{Scale What Counts, Mask What Matters: Evaluating Foundation Models for Zero-Shot Cross-Domain Wi-Fi Sensing}

\begin{abstract}




\noindent
While Wi-Fi sensing offers a compelling, privacy-preserving alternative to cameras, its practical utility has been fundamentally undermined by a lack of robustness across domains. Models trained in one setup fail to generalize to new environments, hardware, or users, a critical "domain shift" problem exacerbated by modest, fragmented public datasets. We shift from this limited paradigm and apply a foundation model approach, leveraging Masked Autoencoding (MAE) style pretraining on the largest and most heterogeneous Wi-Fi CSI datasets collection assembled to date. Our study pretrains and evaluates models on over 1.3 million samples extracted from 14 datasets, collected using 4 distinct devices across the 2.4/5/6 GHz bands and bandwidths from 20 to 160 MHz. Our large-scale evaluation is the first to systematically disentangle the impacts of data diversity versus model capacity on cross-domain performance.  The results establish scaling trends on Wi-Fi CSI sensing. First, our experiments show log-linear improvements in unseen domain performance as the amount of pretraining data increases, suggesting that data scale and diversity are key to domain generalization. Second, based on the current data volume, larger model can only provide marginal gains for cross-domain performance, indicating that data, rather than model capacity, is the current bottleneck for Wi-Fi sensing generalization. Finally, we conduct a series of cross-domain evaluations on human activity recognition, human gesture recognition and user identification tasks. The results show that the large-scale pretraining improves cross-domain accuracy ranging from 2.2\% to 15.7\%, compared to the supervised learning baseline. Overall, our findings provide insightful direction for designing future Wi-Fi sensing systems that can eventually be robust enough for real-world deployment.
\end{abstract}

\acmSubmissionID{217}
\maketitle
\section{INTRODUCTION}

Wireless sensing repurposes communication signals to perceive people and environments without cameras or wearables, offering resilience to occlusion and stronger privacy guarantees than vision. Among radio options, Wi-Fi is especially attractive: it is already deployed at scale indoors, operates in unlicensed bands, and exposes channel measurements on commodity hardware in many settings \cite{nexmon, HernLightweight, halperin2011tool}. Compared with mmWave or UWB, Wi-Fi’s ubiquity and low marginal cost makes it a pragmatic substrate for broad adoption, particularly in homes, offices, and hospitals where installing new infrastructure is costly. These advantages motivate a sensing stack that “rides” existing networks rather than competing with them, setting the stage for methods that generalize across real deployments rather than lab prototypes. In short, Wi-Fi sensing pairs pervasive coverage with low operational friction, making it a practical substrate for ubiquitous perception in real environments.

\begin{figure}[t!]
\vspace{5mm}
    \centering
    \includegraphics[width=\columnwidth]{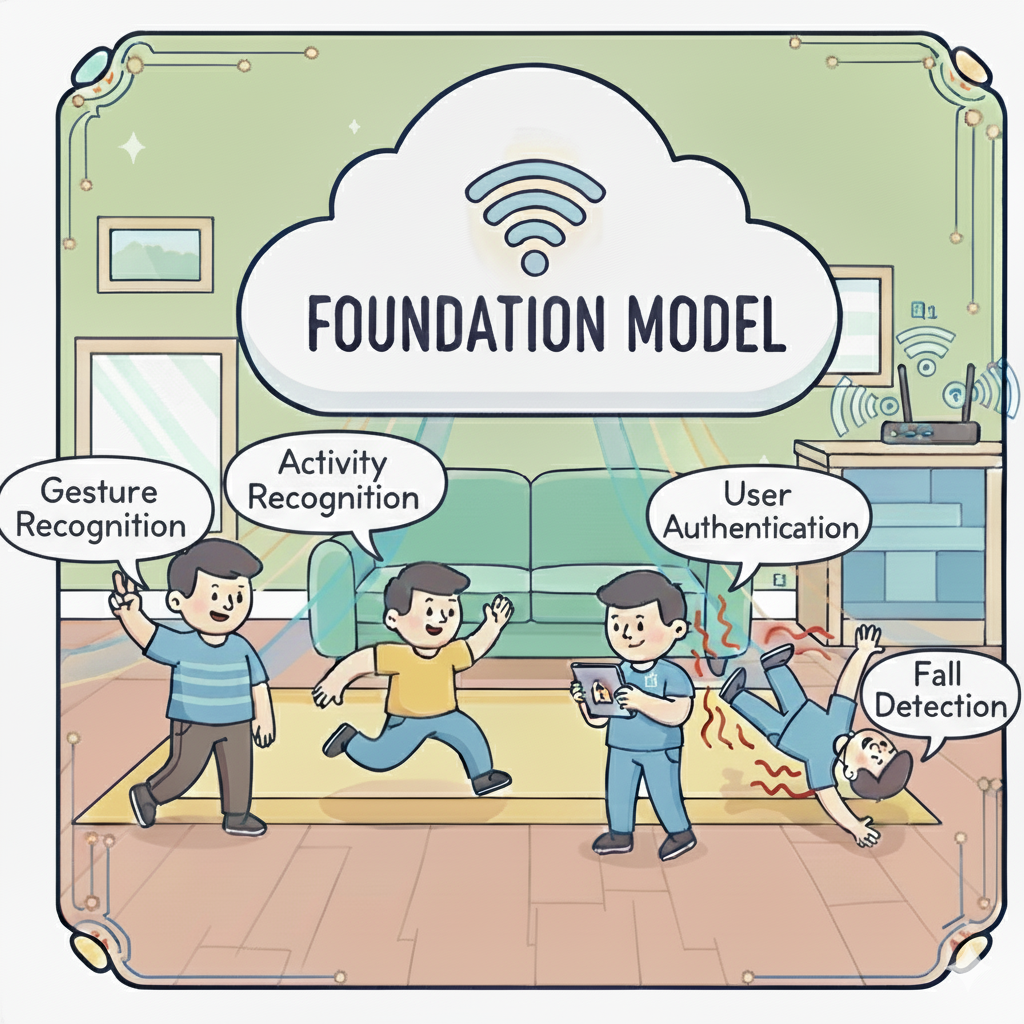}
    \caption{Foundation Model powering various Wi-Fi sensing tasks (Image generated by Google Gemini).}
    \label{fig:wifi_sensing_foundation_model}
\end{figure}

Over the past decade, Wi-Fi sensing has advanced from proof-of-concept demos to systems reporting strong results on activity recognition, vital-sign monitoring, and indoor localization. Yet robustness remains unresolved. Models trained in one room, with one hardware stack or one population, often degrade under “domain shift” when deployed elsewhere. Shifts arise from room geometry, furniture and occupancy changes, antenna placement, bandwidth and center-frequency variation, vendor-specific baseband processing, and time-varying interference. Community data issues compound the problem: compared to vision datasets, public Wi-Fi sensing datasets are modest in size, unevenly annotated, and fragmented by incompatible preprocessing, which hinders comparability and encourages narrow overfitting rather than general solutions. As a result, today’s systems perform well in curated settings but struggle to deliver consistent accuracy across environments, sessions, hardware, and unseen subjects. The enduring challenge is to learn representations that are stable across deployments while retaining sensitivity to the fine temporal cues that drive sensing performance.

Recent progress in computer vision and speech suggests a path forward. Models pretrained on large-scale unlabeled images/audios data using self-supervised and unsupervised methods, such as contrastive learning \cite{chen2020simple, oord2018representation}, masked autoencoding (MAE) \cite{he2022masked}, and diffusion generative pretraining \cite{ho2020denoising}, have produced transferable representations that fine-tune effectively on small labeled datasets. Though Wi-Fi CSI can be converted to heatmaps, a special kind of image, it differs in key ways: signals are complex-valued with both amplitude and phase information; the axes carry physical meanings (time, subcarrier, antenna) rather than Euclidean geometry; and acquisition is tightly coupled to hardware and protocol choices. But a recent study \cite{xu2025evaluating} has demonstrated that the MAE pretraining is still effective when applied to Wi-Fi CSI-based sensing, highlighting its value for various Wi-Fi sensing tasks. However, the current effort is still limited in scope. Pretraining only relies on a few small datasets, and evaluations are confined to within-dataset random split cross-validation or narrow few-shot transfer settings, leaving scalability and cross-domain generalization largely unexplored. We therefore pivot from small, curated corpora to a foundation-model setting and evaluate at \textbf{\emph{scale}} on broad, heterogeneous CSI corpora designed to stress transfer across environment, hardware, bands, and subjects. Below are our summarized contributions:

\begin{itemize}
\item We perform the largest and most comprehensive study of scalability and generalization for Wi-Fi sensing foundation models to date, pretraining and testing on 14 datasets totaling 1320892 CSI samples after data cleaning and preprocessing, collected with 4 distinct Wi-Fi devices across the 2.4/5/6 GHz bands and bandwidths from 20 to 160 MHz, enabling rigorous cross-device, cross-band, cross-subject and cross-site transfer tests.
\item Cross-domain generalisation has always been a pain point for Wi-Fi sensing. Our  results show that the large-scale pretraining improves cross-domain accuracy ranging from 2.2\% to 15.7\%, compared to the supervised learning baseline. 
\item We report scaling trends of Wi-Fi sensing foundation models that disentangle data diversity from model capacity under the cross-domain setting. The results provide insightful direction for future Wi-Fi sensing foundation model designs towards better generalization. For the scaling with respect to data sizes, we observe a linear relationship between accuracy of cross-domain and the logarithm of data size. This shows that data scale helps to improve cross-domain accuracy. 
\item We show that, if we use the maximum data volume, larger models can only provide marginal improvements to cross-domain performance. This indicates that model capacity is the current bottleneck and it may be possible to further improve cross-domain performance if we can scale up the data volume. 
\end{itemize}

\section{PRELIMINARY} \label{prelimnary}

\subsection{Fundamentals of Wi-Fi Sensing}
Wi-Fi sensing treats the wireless channel as a spatiotemporal filter whose complex frequency response encodes motion and geometric properties in the environment. Most Wi-Fi sensing set-ups consider a transmitter with multiple antennas, and one or more receiver each equipped with multiple antennas. Let $N_{\mathrm{tx}}$ be the number of transmitting antennas and $N_{\mathrm{rx}}$ be the number of receiving antenna chains, which is the product of the number of receivers and the number of antennas per receiver. Now, let $H_{ij}(f,t)\in\mathbb{C}$ denote the channel frequency response (CFR) between transmit antenna $j\in\{1,\ldots,N_{\mathrm{tx}}\}$ and receive antenna chain $i\in\{1,\ldots,N_{\mathrm{rx}}\}$ 
at carrier-offset frequency $f$ and slow time scale $t$; the corresponding channel impulse response (CIR) is
\begin{equation}
h_{ij}(\tau,t)=\int H_{ij}(f,t)\,e^{j2\pi f\tau}\,\mathrm{d}f .
\end{equation}
where $\tau$ is the fast time scale. 
A standard multipath model links CFR to propagation. Writing array steering vectors as $\mathbf{a}_r(\theta_p)\in\mathbb{C}^{N_{\mathrm{rx}}}$ and $\mathbf{a}_t(\phi_p)\in\mathbb{C}^{N_{\mathrm{tx}}}$ for angle of arrival $\theta_p$ and departure $\phi_p$, a narrowband path $p$ with delay $\tau_p$, Doppler $\nu_p$, and complex gain $\alpha_p$ contributes
\begin{equation}
\mathbf{H}(f,t)\;=\;\sum_{p=1}^{P}\alpha_p\,e^{-j2\pi f\tau_p}\,e^{j2\pi \nu_p t}\,\mathbf{a}_r(\theta_p)\mathbf{a}_t(\phi_p)^{\mathrm{H}}\;+\;\mathbf{E}(f,t) ,
\end{equation}
where $\mathbf{H}\in\mathbb{C}^{N_{\mathrm{rx}}\times N_{\mathrm{tx}}}$ and $\mathbf{E}$ models estimation noise and hardware distortions. Motion perturbs $\{\alpha_p,\nu_p\}$, while geometry controls $\{\tau_p,\theta_p,\phi_p\}$; hence time--frequency structure in $H$ carries information about dynamics and scene layout. This decomposition motivates feature extractors that respect linear phase with delay, sinusoidal evolution with Doppler, and array-manifold structure across antennas, preparing the ground for invariant representations in the later section.

In OFDM systems, like Wi-Fi, the CFR is sampled on a discrete lattice $f_k=f_0+k\Delta_f$ across $N_f$ subcarriers and over packet times $t_n=n\Delta_t$, yielding measurements $H_{ij}[k,n]$ commonly referred to as channel state information (CSI). Different hardware platforms, PHY/driver stacks, and standard revisions normally yield different levels of granularity, leading to different $N_{\mathrm{tx}}$, $N_{\mathrm{rx}}$, $\Delta_f$, $N_f$, and $\Delta_t$. As a result, these parameters vary across different datasets and cause the heterogeneity of CSI. Let's further form these sample to a data tensor $\mathbf{X}\in\mathbb{C}^{N_t\times N_{\mathrm{rx}}\times N_{\mathrm{tx}}\times N_f}$ with indices $(n,i,j,k)$. For CSI-based Wi-Fi sensing, the objective is to map $\mathbf{X}$ to semantics such as environment status, human activities, motion trajectories, etc.  

\subsection{Efforts Towards Domain Independence}
Wi-Fi sensing models often face severe domain shift, where a model trained in one setting fails in another, since wireless Channel State Information (CSI) captures environment- and device-specific multipath characteristics. In other words, the “domain” (physical environment, user morphology, hardware, etc.) imprints on the CSI, making models inherently domain-specific. Recent research has therefore focused on strategies to achieve domain-independent sensing, ensuring that human activity, gesture, fall, or intrusion detection models maintain accuracy across new environments, users, and devices without extensive re-training. Broadly, solutions fall into two categories: (1) feature engineering that extracts invariant signal features, and (2) model design approaches (incl. training frameworks) that explicitly mitigate or adapt to domain differences.

\subsubsection{Feature Engineering and Signal Processing Approaches}
One approach is to handcraft or pre-process input features so that the influence of any particular domain is minimized. A fundamental technique is to remove the static or baseline wireless channel response of an environment, isolating only the perturbations caused by human movement. By high-pass filtering or subtracting a running average of the CSI, one can cancel out the environment’s static multipath effects, leaving dynamic features that generalize better, such as Doppler Frequency Shift (DFS). For example, conjugate multiplication of CSI from two antennas on the same device can effectively filter out quasi-static offsets and device-specific phase noise. This method is simple and provides immediate generalization benefits by eliminating environment-specific biases. However, it may inadvertently filter out slow-motion or posture-related signal variations and cannot address other domain mismatches (e.g., different devices or subject behaviors), limiting its overall effectiveness.

Another strategy is to design invariant features grounded in physical insights about human motion. A prime example is the Body-coordinate Velocity Profile (BVP) introduced by Widar3.0 \cite{zheng2019zero}. BVP represents the power of motion at various velocities in the person's own coordinate system, essentially decoupling the gesture signature from the absolute radio environment. Theoretically, each gesture type yields a unique velocity pattern regardless of where or how the person is oriented. However, computing BVP is complex and often requires multiple transceivers placed at specific positions to resolve ambiguous velocity components for each axis of body coordinates.

Researchers have also explored transforming CSI into domains where movement-induced patterns are more consistent. For instance, taking the time-frequency spectrogram of CSI and extracting dominant energy or periodicity can capture movement-induced Doppler shifts in a normalized way; similarly, wavelet decompositions can emphasize repeating motion patterns while smoothing out environmental specific distortions \cite{zhang2018crosssense}. However, these transformations introduce trade-offs where some fine-grained details may be lost in the abstraction process, and effective use often requires careful empirical tuning (e.g., selecting appropriate wavelet levels or spectrogram window sizes).

\subsubsection{Data Augmentation- and Generative Adversarial-Based Approaches}
Beyond feature engineering, a complementary strategy is to enrich the training data itself via data augmentation so that models learn to handle domain variability. By synthetically perturbing CSI inputs in domain-relevant ways, augmentation can expose the model to a broader spectrum of channel conditions and hardware differences during training, thereby improving robustness to domain shifts. Common augmentations in wireless sensing include time-series jittering (shifting or slightly warping CSI time indices) to simulate timing misalignment or activity duration variations, and noise injection (adding artificial noise to CSI samples) to mimic measurement noise or interference \cite{zhang2020data}. Recent work also explores adjusting entire CSI spectrograms akin to image augmentations \cite{shorten2019survey}: for instance, circularly shifting the time axis or randomly cropping/resizing spectrogram segments. However, data augmentations on CSI can sometimes produce diminishing returns due to over-distortion or conflicting effects, so careful tuning is needed as well.

Data augmentation can extend beyond simple transformations to adversarial \cite{goodfellow2020generative} or generative augmentation \cite{ho2020denoising}: another class of solutions tackles the domain shift problem at the algorithmic level. One prominent strategy is adversarial domain adaptation, where a model is trained with an auxiliary domain classifier to encourage the extracted features to be indistinguishable between the source and target domains. In practice, this is achieved via an adversarial loss (e.g., a gradient reversal layer) that forces the feature encoder to “fool” the domain discriminator, thereby producing representations that contain task-relevant information (e.g., activity classes) but minimal domain cues (environment, user, etc.). Wi-Fi sensing works also add domain discriminators to activity recognition networks or use GAN-based frameworks to align cross-domain feature distributions \cite{zhang2022metaganfi, zhang2023sida}. However, performance can be brittle: if domain shift is large or alignment is mis-targeted, the model may under-align (residual domain leakage) or over-align (suppressed discriminative features). It also demands sufficient target data and careful tuning to avoid negative transfer.

\subsection{Era of Self-Supervised Learning and Foundation Models}
Modern AI research is now entering an era of foundation models built on the rapid advances in Self-Supervised Learning (SSL). Instead of training task-specific models from scratch, the foundation model paradigm pretrains large models on \textbf{a vast amount of unlabeled data} using self-supervised objectives, yielding \textbf{generalizable representations} that can be adapted across various downstream tasks. Recent advances in SSL from vision and Natural Language Processing (NLP) domains have enabled this shift. For example, Contrastive learning approaches such as SimCLR \cite{chen2020simple} and MoCo \cite{he2020momentum} pioneered instance discrimination, training encoders to maximize agreement between different augmented views of the same input while pushing apart others. Another influential direction is predictive coding. For example, Contrastive Predictive Coding\cite{oord2018representation} uses an autoregressive model to predict future input segments in a latent space, with a contrastive loss that forces the latent representation to capture information maximally useful for prediction.

In parallel, generative self-supervised models have propelled the field toward more expressive foundation models. For example, denoising diffusion models learn to gradually corrupt and then reconstruct data, and have achieved breakthroughs in producing high-fidelity images by reversing a noising process. Later on, Masked Autoencoding (MAE) strategies, inspired by the success of BERT \cite{devlin2019bert} in NLP, have gained prominence in vision and beyond. By masking parts of the input and training the model to predict the missing content, masked modeling forces learning of high-level structure and context. More importantly, MAE pretraining does not require careful positive/negative pair design or large contrastive batches as its learning signal comes from the input itself. This simple yet powerful approach has made MAE a popular choice among SSL methods. Besides, a recent study \cite{xu2025evaluating} systematically evaluated various SSL algorithms for CSI-based Wi-Fi sensing tasks and found that \textsc{MAE} nearly achieves top performance across different settings. However, their evaluation still remains \textbf{small scale} and relying on a few datasets with \textbf{within-dataset splits} or \textbf{narrow few-shot transfers}. Therefore, scalability and broad cross-domain generalization are not well established. This gap motivates the next section (\cref{sec:method}), where we move from method-level SSL comparisons to a \textbf{foundation model} setting that pretrains on large pools of Wi-Fi CSI via the MAE.

\begin{table*}[]
\caption{Public datasets used in Foundation Model pretraining and evaluation.} 
\label{tab:datasets}
\centering
\resizebox{\textwidth}{!}{
    \begin{tabular}{lccccccc}
        \toprule\toprule
        \textbf{Dataset} & \textbf{Frequency Band} & \textbf{Bandwidth(No.Subcarriers)} & \textbf{TX/RX Ants} & \textbf{Sampling Rate} & \textbf{Device}  & \textbf{Tasks} & No.Samples (after prep.)\\ 
        \midrule
        Person-in-WiFi 3D \cite{yan2024person} & 5GHz & 20MHz (30) & 3/3 & 300Hz &Intel 5300  & 3D Pose Estimation & 81526\\
        Exposing the CSI \cite{cominelli2023exposing}& 5GHz & 160MHz (2048) & 1/(3x4) &150Hz &ASUS RT-AX86U  & HAR & 79632\\
        XRF55 \cite{wang2024xrf55} & 5GHz & 20MHz (30) & 1/(3x3) &200Hz  &Intel 5300  & HAR & 501600\\
        XRFV2 \cite{lan2025xrf}& 5GHz & 20MHz (30) & 1/(3x3) &200Hz &Intel 5300  & HAR (Segmentation) & 115430\\
        Wi6CSI \cite{kentridgeai_2025} & 6GHz & 160MHz (2025) & 1/4 &800Hz &PicoScenes  & HAR/People Counting & 38208\\
        Widar 3.0 \cite{zhang2021widar3} & 5GHz & 20MHz (30) & 1/(6x3) &1000Hz &Intel 5300  & HGR & 270998\\
        WiMANS \cite{huang2024wimans} & 2.4/5GHz & 20MHz (30) & 3/3 &1000Hz &Intel 5300  & Multi-User HAR & 33858\\
        FallDar \cite{yang2022rethinking} & 5GHz & 20MHz (30) & 1/3 &1000Hz &Not Specified  & Fall Detection & 1655\\
        GaitID \cite{zhang2020gaitid}& 5GHz & 20MHz (30) & 1/(6x3) &1000Hz &Intel 5300  & User Identification & 22497\\
        Wi-MIR \cite{islam2024wi} & 5GHz & 20MHz (30) & 3/3 &950Hz &Intel 5300  & Multi-User HAR & 54930\\
        Behavior-based User Authentication \cite{shi2017smart} & 5GHz & 20MHz (30) & 3/3 &1000Hz &Intel 5300  & User Identification & 82098\\
        NTU-Fi \cite{yang2023sensefi} & 5GHz & 40MHz (114) & 1/3 &500Hz &TP-Link N750 & HAR & 12240\\
        SignFi \cite{ma2018signfi}& 5GHz & 20MHz (30) & 1/3 &200Hz &Intel 5300  & HGR & 24060\\
        MM-Fi \cite{yang2023mm}& 5GHz & 40MHz (114) & 1/3 &1000Hz &TP-Link N750  & 3D Pose Estimation & 2160\\
        \midrule
        Total &  &   &  & & &  &  1320892\\
        \bottomrule
    \end{tabular}
}
\end{table*}

\section{Foundation Models for Wi-Fi Sensing} \label{sec:method}
Building on the motivation for self-supervised pretraining and the emergence of MAE as a stable, scalable objective, we now define what a foundation model for Wi-Fi sensing entails and how we instantiate it \textbf{at scale}.

\begin{figure*}[t!]
    \centering
    \includegraphics[width=\linewidth]{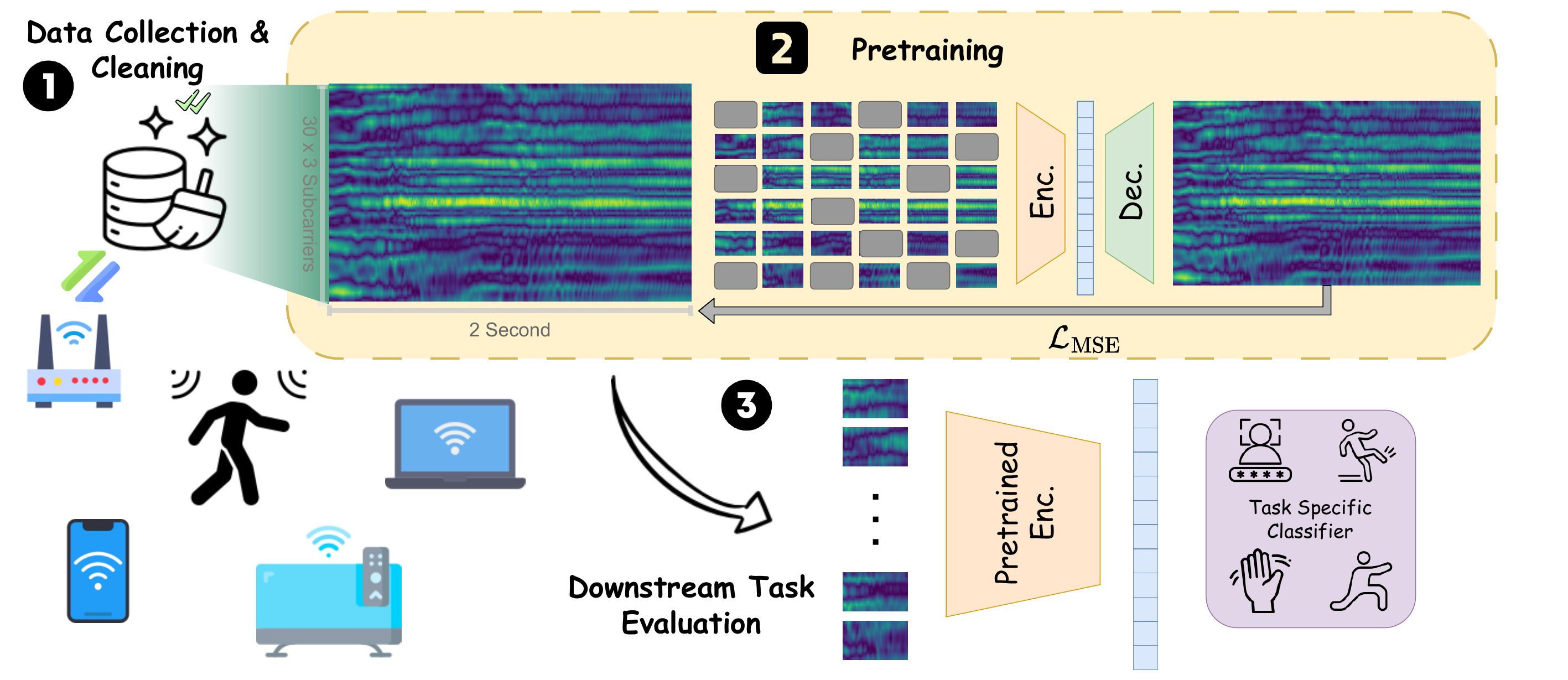}
    \caption{Overview of our Wi-Fi sensing foundation model evaluation. We collect a large-scale and diverse set of Wi-Fi CSI from public datasets that cover various tasks, devices, frequency bands, and environments. After harmonizing and preprocessing the data, we pretrain a Masked Autoencoder (MAE) model on the unified CSI representation. The pretrained encoder can then be fine-tuned on downstream sensing tasks with limited labels.}
    \label{fig:foundation_model_overview}
\end{figure*}

\subsection{Wi-Fi CSI Dataset Collection}

We gather a comprehensive collection that consists of a wide range of publicly accessible Wi-Fi CSI datasets, encompassing both published and unpublished datasets, containing either raw or preprocessed formats, to study the scalability and generalization of the Wi-Fi sensing foundation model. These datasets cover various common sensing tasks such as gesture recognition, activity recognition, user authentication, fall detection, etc. Furthermore, as some collected datasets were captured by different hardware or through different standards, they also provide us with great opportunities to study more cross-domain settings, such as cross-devices and cross-bands. Table.\ref{tab:datasets} summarizes some key attributes of these datasets, and we provide more details for each dataset below:

\noindent\textbf{Person-in-WiFi 3D.} This dataset focuses on multi-person 3D human pose estimation using Wi-Fi CSI captured with Intel 5300 in the 5 GHz band and synchronized with the Azure Kinect RGB-D camera. The dataset collected samples from 7 subjects performing 8 daily actions in 3 different locations.

\noindent\textbf{Exposing the CSI.} 
This dataset was introduced to explore the sensing capabilities of modern Wi-Fi 6 standards. Wi-Fi 6 offers significantly wider bandwidths as well as finer-grained subcarrier resolution compared to previous generations, which can potentially enhance sensing performance. This dataset involves in 3 individual subjects performing 12 different activities in 3 different environments.


\noindent\textbf{XRF55.}
A comprehensive dataset for human activity recognition (HAR) via RF signals, including RFID, Wi-Fi, and mmWave. The ground truth labels were also collected and annotated via synchronized video recordings from Azure Kinect cameras. It is a quite large-scale and challenging dataset that involves 39 subjects performing 55 different activities in 4 different scenes. For the Wi-Fi CSI part, it was also collected by Intel 5300 NICs in the 5 GHz band with standard 20 MHz channels.


\noindent\textbf{XRFV2.} Building upon XRF55, this enhanced version maintains the same hardware setup, but introduces temporal segmentation annotations for continuous activity streams. Rather than isolated gesture instances, it supports research on activity boundary detection and sequential action understanding, which are critical for practical monitoring systems. This dataset involves 15 subjects performing 30 different continuous indoor activities across 3 different environments.

\noindent\textbf{Wi6CSI.} Captured using PicoScenes-enabled devices in the 6 GHz band, this dataset represents the latest Wi-Fi 6E standard with 160 MHz bandwidth. It covers 3 typical sensing tasks: single/multi-person human activity recognition and people counting. For single-person activity recognition, it involves 5 subjects performing 13 different activities in 3 different rooms. The multi-person activity recognition task includes 3 subjects performing the same 13 activities together, while the people counting task involves varying numbers of people (1-5) in a room while the subjects either sit still or stand and mingle with other paticipants. Note that there is no publication associated with this dataset yet. We found the dataset and its details from the huggingface repository\footnote{https://huggingface.co/datasets/kentridgeai/Wi6CSI} released by kentridgeai.

\noindent\textbf{Widar 3.0.} Designed for hand gesture recognition, Widar 3.0 uses Intel 5300 hardware in the 5 GHz band at 20 MHz bandwidth. Its distinctive configuration with 1 transmit antenna and 18 receive antennas (6 receivers with 3 antennas each) sampled at 1000 Hz. The dataset includes extensive cross-domain splits across 3 environments, 5 orientations, and 16 users performing 22 gestures, making it a representative benchmark for gesture recognition robustness.

\noindent\textbf{WiMANS.} This another multi-user human activity recognition dataset operates in both 2.4 GHz and 5 GHz bands with 20 MHz bandwidth using Intel 5300 NICs. The dual-band coverage and multi-user scenarios make it valuable for studying cross-band transfer and complex activity disentanglement. This dataset involves 6 subjects performing 9 different activities in 3 different rooms.

\noindent\textbf{FallDar.} Focused on fall detection for elderly care applications, this dataset also uses Intel 5300 in 5 GHz at 20 MHz bandwidth. Sampled at 1000 Hz, it captures both fall events and activities of daily living, providing critical data for safety monitoring systems. The dataset includes 5 subjects and 1 bionic bot performing scripted falls along with normal activities in 3 scenes. In total, the dataset provides 606 fall instances and 1049 non-fall instances. Note that this dataset was accquired from the SDP platform \footnote{http://www.sdp8.net/Dataset?id=dfe4621e-4774-4ef4-b064-e4a082c12335}.

\noindent\textbf{GaitID.} This dataset targets user identification through gait patterns captured via Wi-Fi CSI using Intel 5300 hardware in the 5 GHz band at 20 MHz bandwidth and 1000 Hz sampling. This dataset enables research on biometric authentication without requiring cameras or cooperation from subjects. It involves 11 subjects walking naturally following 4 tracks in each of 3 different environments, providing multiple sessions per user for robust identification. Note that this dataset was accquired from the SDP platform \footnote{http://www.sdp8.net/Dataset?id=87a65da2-18cb-4b8f-a1ec-c9696890172b}.

\noindent\textbf{Wi-MIR.} A multi-user interaction activity recognition dataset, Wi-MIR also employs Intel 5300 NICs operating at 5 GHz with 20 MHz bandwidth. It involves 16 pairs of subjects performing 17 interaction activities in a furnished room.

\noindent\textbf{Behavior-based User Authentication.} This dataset explores user identification through behavioral patterns in CSI. Operating at 5 GHz with 20 MHz bandwidth, it also Intel 5300 hardware. Rather than explicit gestures, it captures ambient behavioral signatures during normal daily activities, enabling passive authentication research for security applications. This dataset contains 8 walking and 8 stationary activities performed by 11 and 5 subjects, respectively, in 2 different environments.


\noindent\textbf{NTU-Fi.} Instead of using Intel 5300, this dataset contains CSI amplitude captured using TP-Link N750 commercial, which operates in the 5 GHz band at 40 MHz bandwidth, providing 114 subcarriers. The dataset contains two sensing tasks, human activity recognition (HAR) and user identification tasks. For the HAR task, it involves 6 different activities, and for the user identification task, it involves 14 different subjects.

\noindent\textbf{SignFi.} A hand gesture recognition dataset using Intel 5300 hardware in the 5 GHz band at 20 MHz bandwidth. The dataset provides 5520 CSI clips for 276 signs (lab setting) and 2760 clips for the same 276 signs (home setting), as well as an additional 7500 clips of 150 signs performed by 5 different users to evaluate user independence.

\noindent\textbf{MM-Fi.} This multimodal dataset combines Wi-Fi CSI with synchronized video for 3D pose estimation. Using TP-Link N750 routers at 5 GHz with 40 MHz bandwidth (114 subcarriers). The paired modalities enable research on cross-modal learning and provide ground-truth validation for pose estimation algorithms, bridging radio-based sensing with established vision benchmarks. It includes 27 categories of daily and rehabilitation actions performed by 40 participants, with data captured concurrently from five modalities: RGB video, depth (stereo IR) frames, LiDAR point clouds, mmWave radar, and Wi-Fi CSI.

\subsection{Data Preprocessing and Cleaning}

Our dataset collection contains heterogeneous CSI data collected from diverse devices, frequency bands, sampling rate and antenna configurations. To make pretraining stable and fair across datasets, we perform several preprocessing and cleaning steps that preserves physically meaningful structure while removing artifacts caused by noise and hardware impairments.

\subsubsection{Harmonizing heterogeneous CSI datasets}
Different datasets expose measurements in incompatible formats: some provide the raw complex channel coefficients, while others release only preprocessed amplitude traces. To accommodate all these datasets into a common pretraining pipeline, we first convert every CSI sample to a real-valued amplitude representation. Concretely, for any complex channel matrix $\mathbf{H}$, we define $\mathbf{X}=|\mathbf{H}|\in\mathbb{R}^{N_t\times N_{\mathrm{rx}}\times N_{\mathrm{tx}}\times N_f}$, which removes phase and yields a consistent amplitude-based real tensor across datasets. \textbf{Note that this is the most generic representation of Wi-Fi CSI used in existing literature, thereby allowing us to incorporate the maximum number of datasets for pretraining.}
Beyond representation, datasets differ in temporal duration $N_t$, subcarrier counts $N_f$, and antenna layout $N_{\mathrm{rx}}$ and $N_{\mathrm{tx}}$. Here $N_{\mathrm{rx}}$ denotes the total number of receive antenna  chains, which is equal to the product 
$N_{\mathrm{recv}} \times N_{\mathrm{apr}}$ where $N_{\mathrm{recv}}$ is the number of receivers and $N_{\mathrm{apr}}$ is the number of antennas per receiver  . From Table.~\ref{tab:datasets}, the most prevalent configuration is a single transmit antenna and one receiver with three receive antennas. We therefore adopt this as our canonical layout and rearrange all antenna configurations to match it by treating multiple transmit antennas and multiple receivers as separate instances in the batch. Table.~\ref{tab:datasets} shows that some receivers have 4 antennas, such as commercial APs or PicoScenes devices, we only use the first three antennas for each of these receivers. 

Formally, for a sample with original tensor $\mathbf{X}\in\mathbb{R}^{N_t\times N_{\mathrm{rx}}\times N_{\mathrm{tx}}\times N_f}$, we extract a per-link tensor
$
\mathbf{X}_{i,u}=\mathbf{X}[:,\mathcal{R}_u,i,:]\in\mathbb{R}^{N_t\times 3\times 1\times N_f}$
where each $i\in\{1,\ldots,N_{\mathrm{tx}}\}$ corresponds to a transmit antenna and each $u\in\{1,\ldots,N_{\mathrm{recv}}\}$ corresponds to a receiver. For each receiver, we will use 3 receive antennas whose indices are elements of the set $\mathcal{R}_u\subset\{1,\ldots,N_{\mathrm{rx}}\}$. Specifically, $\mathcal{R}_u$ contains the indices $(u-1) N_{\mathrm{apr}} + 1$ to $(u-1) N_{\mathrm{apr}} + 3$ so each $\mathcal{R}_u$ has 3 indices corresponding to 3 receive antennas. To align CSI samples on the time domain, we apply a two-second sliding window with one-second stride to segment each CSI recording into multiple clips. The length of each clip is corresponding to the sampling rate of each dataset, which varies between 200 to 2000 packets. To ensure consistent temporal length across datasets, we resample all clips to a target length $N_t^{\star}=600$ via linear interpolation. Similarly, we unify the subcarrier dimension, first, by segmenting the bandwidth of each CSI clip into multiple 20 MHz channels (as this is the minimal and default bandwidth in most datasets), and then resampling the subcarrier axis of each channel to a fixed number of frequency bins $N_f^{\star}=30$. Therefore, after unifying the antenna layout, time length, and frequency bins, each CSI clip is represented as a tensor $\mathbf{X}_{i,u}\in\mathbb{R}^{600\times 3\times 1\times 30}$. For each CSI clip instance generated from the same original recording, they share the same label (e.g. user, date, environment and class) if available. We squeeze the singleton transmit antenna dimension and flatten antenna dimension into the channel dimension to yield the final input tensor shape $\mathbb{R}^{600 \times 90}$ for pretraining. Finally, for each CSI clip instance, we apply Z-score normalization along the antenna-frequency dimension for each timestamp to avoid impacts of Auto Gain Control (AGC) and stabilize the scale across different devices and environments.

\subsubsection{Sanity Checks and Artifact Removal}

\begin{figure}
    \centering
    \begin{subfigure}{0.48\columnwidth}
        \centering
        \includegraphics[width=\textwidth,keepaspectratio]{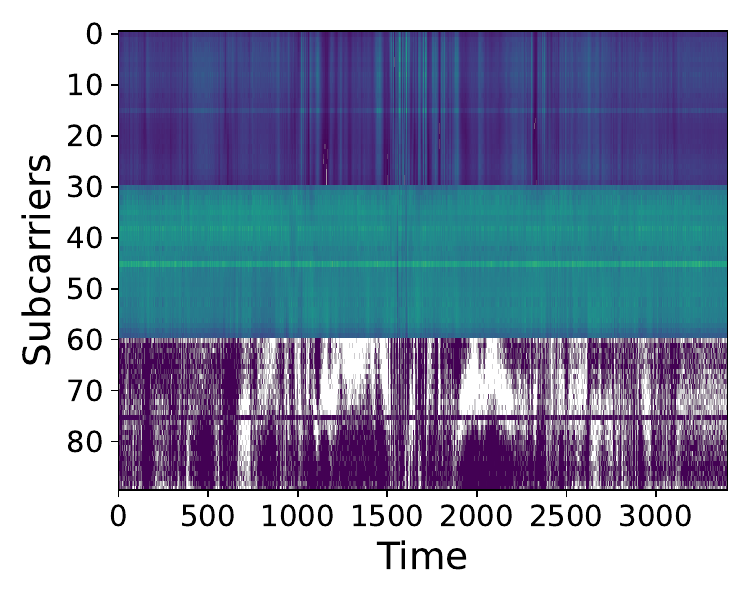}
        \caption{Antenna impairment}
    \end{subfigure}
    \begin{subfigure}{0.48\columnwidth}
        \centering
        \includegraphics[width=\textwidth,keepaspectratio]{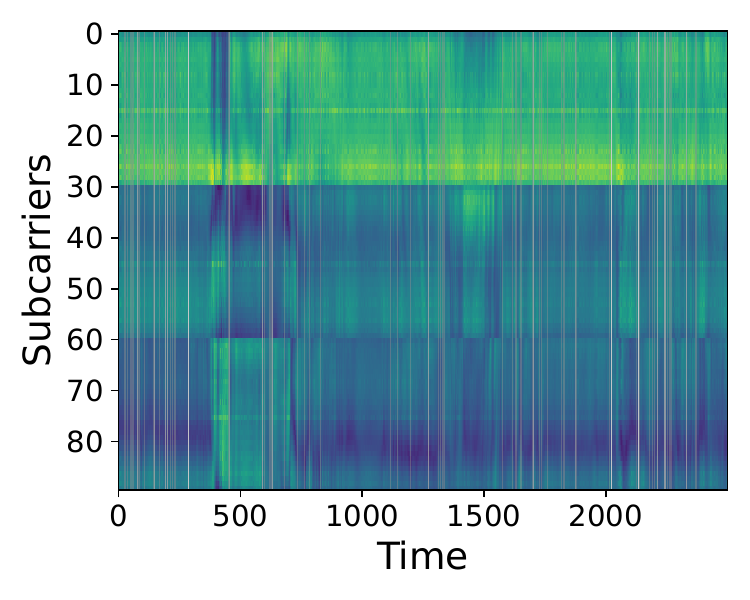}
        \caption{Missing Packets}

    \end{subfigure}
    \begin{subfigure}{0.48\columnwidth}
        \centering
        \includegraphics[width=\textwidth,keepaspectratio]{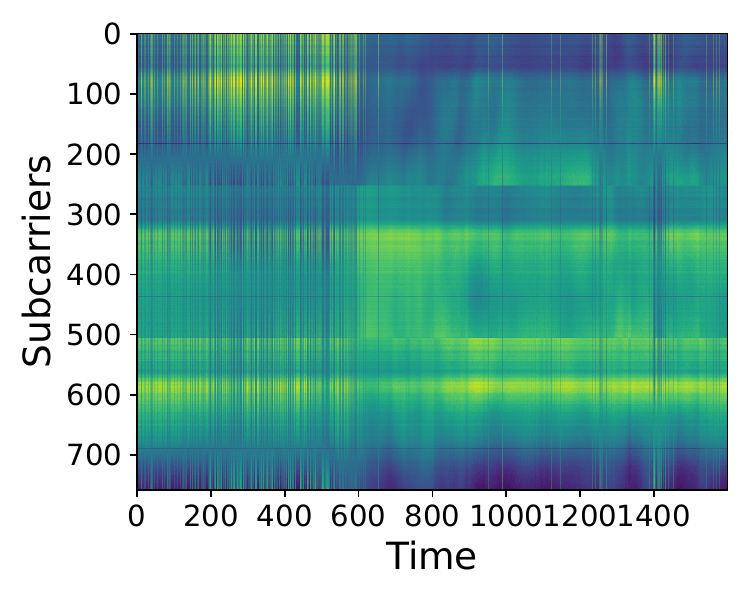}
        \caption{Noisy Packets}
    \end{subfigure}
    \begin{subfigure}{0.48\columnwidth}
        \centering
        \includegraphics[width=\textwidth,keepaspectratio]{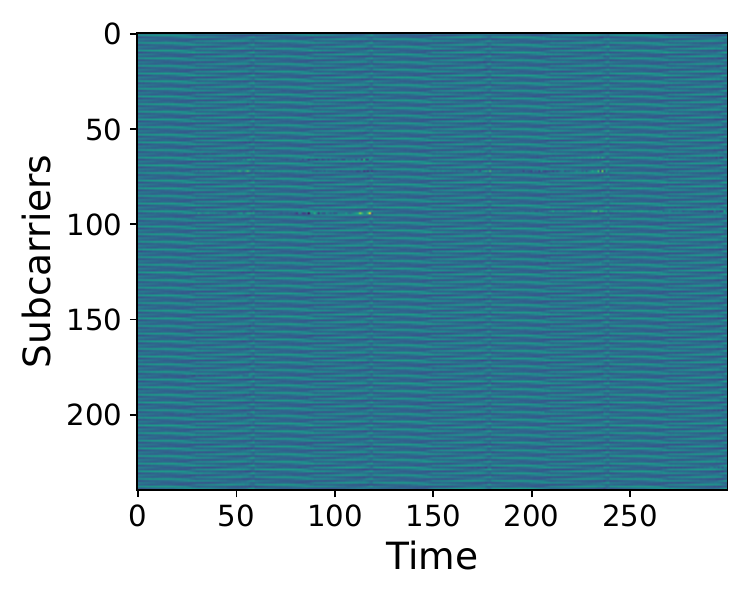}
        \caption{Irregular CSI}
    \end{subfigure}
    \caption{illustration of typical artifacts in CSI data}
    \label{fig:csi_artifacts}
\end{figure}

MAE pretraining relies on the reconstruction loss signal from masked inputs, so it is important to ensure that the data used as the input of pretraining is clean and free from artifacts that could mislead the model or destabilize the training. Though these datasets were collected with care, they may still contain occasional noise spikes, missing packets, or hardware failures. Therefore, in addition to human efforts, we implement a light preprocessing pipeline to enhance data quality. Figure~\ref{fig:csi_artifacts} illustrates some typical artifacts observed in CSI data. The easiest to identify are missing packets, which manifest as all null columns in the CSI tensor. We remove any clips with more than 10\% missing packets. Similarly, antenna impairments can be detected when one or more antennas show abnormally low variance, constant readings or contains irregular null values, indicating hardware failure; such clips are discarded entirely. Next, we detect and remove noisy packets that exhibit extreme amplitude values or sudden jumps compared to neighboring samples ((e.g., exceeding $\mu \pm 2\sigma$)). We replace these outliers with interpolated values from adjacent timestamps. Finally, we screen for irregular CSI patterns that deviate significantly from expected temporal dynamics. Luckily, we find that this type of artifact is grouped inside certain datasets and devices, making it easier to identify and exclude the entire affected records manually. We also report the number of samples after preprocessing in Table~\ref{tab:datasets}.

\subsection{Model Details}

\subsubsection{Architecture Design}
We implement our Wi-Fi foundation model based on the standard Vision Transformer (ViT) architecture~\cite{dosovitskiy2020image} adapted for 2D time-frequency CSI representations. The model follows the Masked Autoencoder (MAE) framework~\cite{he2022masked}, consisting of an asymmetric encoder-decoder structure optimized for CSI reconstruction.

\paragraph{Encoder.} The encoder employs a standard ViT architecture that processes visible (unmasked) patches. After patch embedding via a linear projection layer, learnable positional embeddings are added to retain spatial-temporal awareness. The embedded patches are then fed through a stack of transformer blocks, each containing multi-head self-attention (MHSA) and feed-forward network (FFN) with LayerNorm and residual connections:
\begin{equation}
\begin{aligned}
\mathbf{z}'_\ell &= \mathrm{MHSA}(\mathrm{LN}(\mathbf{z}_{\ell-1})) + \mathbf{z}_{\ell-1}, \\
\mathbf{z}_\ell &= \mathrm{FFN}(\mathrm{LN}(\mathbf{z}'_\ell)) + \mathbf{z}'_\ell,
\end{aligned}
\end{equation}
where $\mathbf{z}_0$ denotes the embedded visible patches and $\ell\in\{1,\ldots,L_{\mathrm{enc}}\}$ indexes encoder layers. The FFN consists of two linear layers with GELU activation and an expansion ratio of 4. By processing only visible patches, the encoder achieves significant computational savings, which is crucial for large scale pretraining.

\paragraph{Decoder.} The decoder is a lightweight transformer that reconstructs the full CSI clip from the encoded visible patches plus learnable mask tokens. A linear projection first maps encoder outputs to the decoder's hidden dimension, then mask tokens are appended at the positions of removed patches. Positional embeddings (shared with the encoder) are added before passing through $L_{\mathrm{dec}}$ transformer blocks with the same MHSA-FFN structure. Finally, a linear projection head predicts the pixel values for each patch. The asymmetric design, where the decoder is narrower and shallower than the encoder, reduces pretraining cost while enforcing encoder to learn richer representations.

\paragraph{Patch Embedding and Tokenization.} Given an input CSI clip $\mathbf{X}\in\mathbb{R}^{N_t\times N_c}$ with $N_t=600$ timestamps and $N_c=90$ antenna-frequency channels, we partition it into non-overlapping 2D patches of size $P_t\times P_f$ along the time and channel dimensions. Each patch is flattened and linearly projected to a $D$-dimensional token (where $D$ is the hidden dimension). For example, if the patch size is $(P_t, P_f)=(30, 3)$, the input is divided into $N_p = \frac{600}{30}\times\frac{90}{3}=600$ patches. Learnable 1D positional embeddings are added to preserve patch order, encoding both temporal progression and spatial structure across antennas and subcarriers.

\subsubsection{Masking Strategy}
During pretraining, we randomly mask a high proportion $\rho$ of the patches and reconstruct them. Specifically, we sample a binary mask $\mathcal{M}\in\{0,1\}^{N_p}$ where $\mathcal{M}_i=0$ indicates that patch $i$ is masked. Only visible patches ($\mathcal{M}_i=1$) are encoded, significantly reducing computation. The decoder receives the encoded visible patches along with learnable mask tokens at masked positions, then predicts the original pixel values for all patches. Following MAE, we compute the reconstruction loss only on masked patches:
\begin{equation}
\mathcal{L}_{\mathrm{MSE}} = \frac{1}{|\{i:\mathcal{M}_i=0\}|} \sum_{i:\mathcal{M}_i=0} \|\hat{\mathbf{x}}_i - \mathbf{x}_i\|_2^2,
\end{equation}
where $\mathbf{x}_i$ is the original patch and $\hat{\mathbf{x}}_i$ is the reconstructed patch. Random masking forces the encoder to learn global context and temporal-spectral correlations from limited observations, producing transferable representations.

\subsubsection{Training Objective and Implementation}
The pretraining objective is to minimize the $\mathcal{L}_{\mathrm{MSE}}$ between reconstructed and original masked patches. We implement the model in PyTorch and leverage mixed-precision training (BF16) with automatic gradient scaling to accelerate training on modern GPUs. After pretraining, the encoder serves as a feature extractor for downstream sensing tasks. We discard the decoder and attach a lightweight task-specific head (an MLP layer) to the encoder's [CLS] token.

\section{EVALUATION} \label{sec:evaluation}

We conduct extensive experiments to evaluate the scalability and generalization capabilities of Wi-Fi sensing foundation models. Our evaluation addresses three key questions: \textbf{(1) How does model capacity influence generalization under domain shift? (2) How do pretraining data scale and diversity affect cross-domain transfer performance? (3) How do different architectural and training choices impact robustness across heterogeneous sensing scenarios?} We organize our experiments into systematic studies of data scaling, model scaling inline with cross-domain transfer tests that stress generalization across environments, hardware, frequency bands, and subjects.

\subsection{Experimental Setup}

\subsubsection{Pretraining Configuration} \label{sec:pretraining_setup}
We pretrain our foundation models on the collected datasets described in Table~\ref{tab:datasets}, using the MAE framework with masking ratio $\rho=80\%$ and patch size (30,3). We justify these design choices in \cref{sec:mask_ratio} and \cref{sec:patch_size}. The default encoder configuration uses the ViT-Small variant with 8 layers, 384 hidden dimensions, and 6 attention heads, along with a shallow decoder that uses 4 layers with 512 hidden dimension and 8 attention heads. We train with AdamW optimizer, learning rate $1e^{-4}$ with cosine decay and 1000 steps warmup, batch size 128, and weight decay 0.03. All models are trained on NVIDIA RTX 4090/H100 GPUs using Distributed Data Parallel (DDP) with mixed precision. Training employs early stopping after 5 epochs without validation improvement.

\subsubsection{Downstream Tasks and Evaluation Protocol} \label{sec:downstream_setup}
We evaluate the performance across three representative sensing tasks on a few published datasets: (1) Human Activity Recognition (HAR) on WiMANS and XRF55, (2) Hand Gesture Recognition (HGR) on Widar 3.0, and (3) User Identification on GaitID. We report accuracy for classification tasks. We consider two data splitting methods:
\begin{enumerate}
    \item  We use standard 80-20 \textbf{in-domain} cross-validation splits (based on repetitions if applicable, otherwise random-based);   
    \item and to assess \textbf{cross-domain} generalization, we employ leave-one-domain-out evaluation where models trained on source domains are tested on unseen target domains differing in environment or subject.
\end{enumerate}

Note that due to the fragmentation of Widar 3.0 dataset, only part of dataset are used for evaluation. To be more specific, we use "\verb|CSI/20181130|" for cross-subject evaluation, which consists of 9 subjects performing 9 gestures in a single room. To study the cross-environment generalization, we combine "\verb|CSI/20181109|", "\verb|CSI/20181115|", ""\verb|CSI/20181118|", "\verb|CSI/20181121|", "\verb|CSI/201812|\\\verb|04|", "\verb|CSI/20181205|" and "\verb|CSI/20181208|" to form a dataset that contains 3 different subjects performing the same 9 gestures in 2 rooms for cross-environment evaluation. Since these two sets shares the same gestures, we combine both for the cross-validation. Unless otherwise specified, for each evaluation, we first pretrain the model on the CSI data that excludes the testing set (training set + other datasets). After pretraining, we fine-tune the pretrained encoder with a lightweight classification head using the training set via the cross-entropy loss. We use the same optimizer configuration as for pretraining, except the batch size is set to 32.

\subsubsection{Generative Pretraining Results}

\begin{figure*}[t!]
    \centering
    \includegraphics[width=\linewidth]{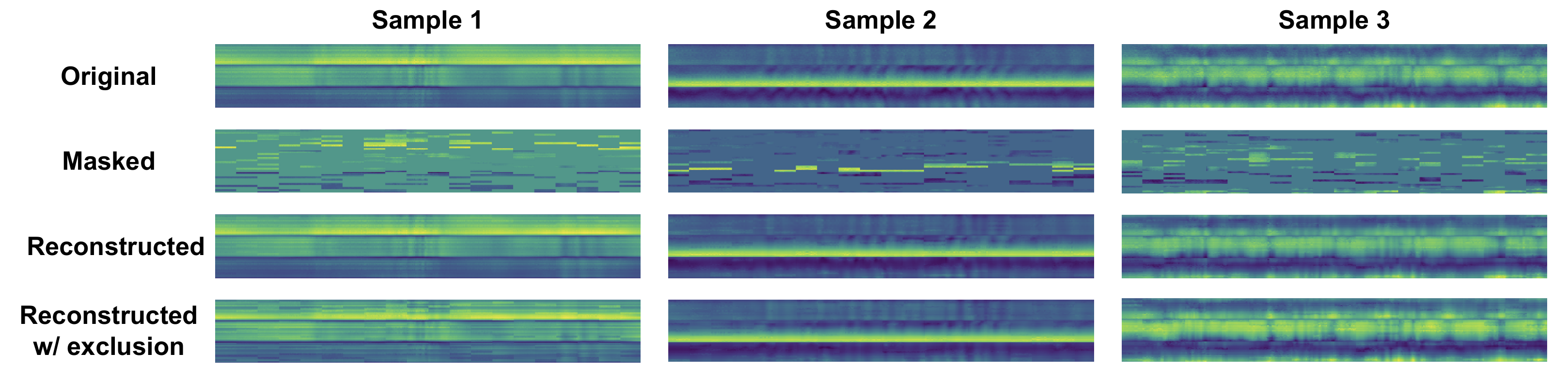}
    \caption{Visualization of original and reconstructed CSI samples. Despite heavy masking, the model effectively reconstructs the missing patches, capturing both temporal dynamics and frequency patterns essential for Wi-Fi sensing tasks.}
    \label{fig:generative_vis}
\end{figure*}

Figure.\ref{fig:generative_vis} visualizes original and reconstructed CSI samples from unseen testing datasets after MAE pretraining with 80\% masking. Despite the aggressive masking strategy that removes the majority of input patches, the model demonstrates remarkable reconstruction fidelity across spatiotemporal dimensions. The reconstructions preserve both sharp temporal transitions corresponding to human motions, as well as frequency-domain fading profile variations caused by human movements. This suggests that the encoder learns to capture global spatiotemporal dependencies rather than merely interpolating local neighbors. Samples from left to right show increasing motion complexity (stronger CSI variations), yet the model consistently recovers key patterns. These results validate the effectiveness of MAE pretraining in learning rich representations from sparse CSI observations, which can be leveraged for downstream sensing tasks.

\subsubsection{Performance on Downstream Tasks}

\begin{table}[]
    \centering  
    \caption{Cross-validation performance on downstream tasks for in-domain data splitting}
    \resizebox{\columnwidth}{!}{
    \begin{tabular}{@{}ccccc@{}}
    \toprule\toprule
    \multicolumn{1}{c|}{Method}   & WiMANS & XRF55 & Widar 3.0 & GaitID \\ \midrule
    \multicolumn{1}{c|}{Supervised}      &    47.8\%    &    76.4\%   &         91.5\%          &      96.2\%     \\ \midrule
    \multicolumn{1}{c|}{Pretrained w/ LP} &   28.3\%     &     70.9\% &     83.2\%    &  95.2\%   \\ \midrule
    \rowcolor{black!8}
    \multicolumn{1}{c|}{Pretrained w/ FT}      &   42.5\% (5.3\%$\downarrow$)  &    77.6\%(1.2\%$\uparrow$)     & 91.5\% (\textemdash) &    98.6\% (2.4\%$\uparrow$)    \\ \bottomrule
    \end{tabular}
    }
    \label{tab:cross_eval}
    \vspace{0mm}
\end{table}

\begin{table}[t]
\centering
\caption{Zero-shot cross-domain performance on downstream tasks}
\label{tab:downstream_performance}
\begin{subtable}{\columnwidth}
    \centering
    \caption{Human Activity Recognition (HAR) results on WiMANS and XRF55 datasets under cross-subject and cross-environment evaluation protocols.}
    \label{tab:har_acc}
    \resizebox{\textwidth}{!}{
        \begin{tabular}{@{} c c c c c c @{}}
            \toprule\toprule
            \multirow{2}{*}{} & \multirow{2}{*}{Method} &\multicolumn{2}{c}{WiMANS} &\multicolumn{2}{c}{XRF55} \\
            \cmidrule(lr){3-4}\cmidrule(lr){5-6}
            & & {cross-subject} & {cross-env} & {cross-subject} & {cross-env} \\
            \midrule
            & Supervised & 39.4\% & 37.7\% & 18.5\% & 6.8\% \\
            \midrule
            & Pretrained w/ LP & 24.4\% & 21.0\% & 26.7\% & 13.2\% \\
            \midrule
            \rowcolor{black!8}
            & Pretrained w/ FT  & 45.7\% (6.3\% $\uparrow$) & 47.3\% (9.6\% $\uparrow$) & 31.1\% (12.6\% $\uparrow$) & 19.8\% (13\% $\uparrow$) \\
            \bottomrule
        \end{tabular}
    }
    \vspace{5mm}
\end{subtable}
\begin{subtable}{\columnwidth}
    \centering
    \caption{Hand Gesture Recognition (HGR) and User Identification results on Widar3.0 and GaitID datasets}
    \label{tab:hgr_ui_acc}
    \resizebox{\textwidth}{!}{
        \begin{tabular}{@{} c c c c c c@{}}
            \toprule\toprule
            \multirow{2}{*}{} & \multirow{2}{*}{Method} &\multicolumn{2}{c}{Widar3.0} &\multicolumn{2}{c}{GaitID} \\
            \cmidrule(lr){3-4}\cmidrule(lr){5-6}
            & & {cross-subject} & {cross-env} & {cross-track} & {cross-env} \\
            \midrule
            & Supervised & 54.7\% & 33.5\% & 92.7\% & 50.4\% \\
            \midrule
            & Pretrained w/ LP & 46.1\% & 29.3\% & 86.1\% & 51.1\% \\
            \midrule
            \rowcolor{black!8}
            & Pretrained w/ FT  & 67.5\% (12.8\% $\uparrow$) & 41.2\% (7.7\% $\uparrow$) & 94.9\% (2.2\% $\uparrow$) & 66.1\% (15.7\% $\uparrow$) \\
            \bottomrule
        \end{tabular}
    }
\end{subtable}
\vspace{-5mm}
\end{table}
We follow the steps in \cref{sec:pretraining_setup} and \cref{sec:downstream_setup} to pretrain the foundation model and evaluate its performance on downstream tasks, such as human activity recognition, hand gesture recognition, and user identification. Additionally, we compare the performance of full fine-tuning (FT) the pretrained model with training a supervised model from scratch as the baseline. We also include the results of linear probing (LP) the pretrained model, where we freeze the encoder and only train a linear classification head on top. Table.\ref{tab:cross_eval} provides the in-domain cross-validation results on representative downstream tasks. We observe that fine-tuning the pretrained model generally matches or slightly underperforms training from scratch, while linear probing underperforms significantly. This suggests that while the learned representations are useful, they still require adaptation to the specific downstream tasks. 

Cross-domain scenarios are the ample scope where foundation models can show their advantages. Table~\ref{tab:downstream_performance} summarizes the zero-shot cross-domain performance of our pretrained foundation models on the same tasks and datasets as in-domain experiments. Compared to supervised baselines trained from scratch, the pretrained models with full fine-tuning achieve significant accuracy improvements (ranging from 2.2\% to 15.7\%) across all tasks and evaluation protocols. Besides, we observe that no matter which downstream task is evaluated, both supervised and pretrained models suffer from cross-environment generalization more than cross-subject generalization, indicating that environmental variations pose a greater challenge for Wi-Fi sensing. The pretrained models help mitigate this gap to some extent.

\subsection{Model Scaling}

We investigate how architectural design choices affect cross-domain transfer performance by systematically varying encoder capacity, masking ratio, and spatiotemporal patch size. To isolate these effects, we conduct controlled experiments on hand gesture recognition using the Widar 3.0 dataset with a cross-domain evaluation protocol.

\subsubsection{Impact of Encoder Capacity}\label{sec:model_capacity}

\begin{figure}[t!]
    \centering
    \begin{subfigure}{\columnwidth}
        \includegraphics[width=\linewidth]{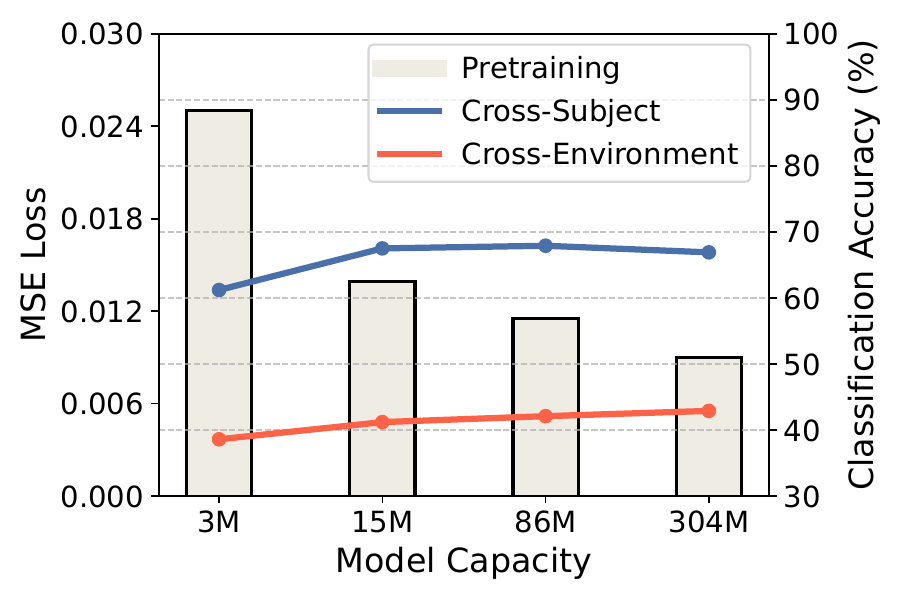}
        \caption{Impact of encoder capacity}
        \label{fig:model_capacities}
    \end{subfigure}

    \begin{subfigure}{\columnwidth}
        \includegraphics[width=\linewidth]{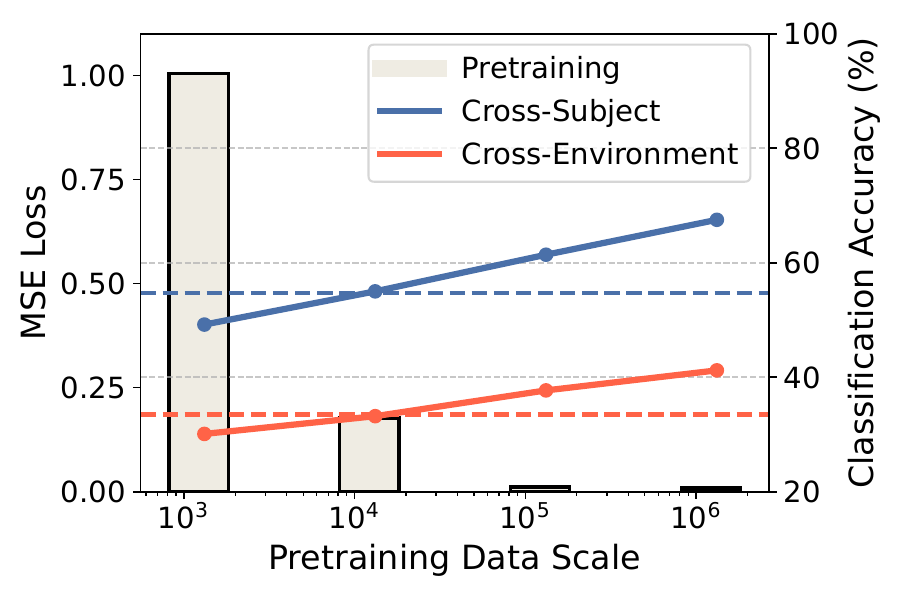}
        \caption{Impact of data scaling. The horizontal dashed lines show the performance of supervised learning in Table \ref{tab:hgr_ui_acc}.}
        \label{fig:data_scaling}
    \end{subfigure}

    \caption{Scaling behaviors of Wi-Fi sensing foundation models under cross-domain evaluation on Widar 3.0 dataset.}
\end{figure}



We evaluate the performance of several well-known ViT variants to study the scaling behavior of foundation models on model sizes for Wi-Fi sensing:
\begin{itemize}
\item \textbf{ViT-Tiny}: $L_{\mathrm{enc}}=6$ layers, hidden dimension $D=192$, 3 attention heads, $\sim$3M parameters.
\item \textbf{ViT-Small}: $L_{\mathrm{enc}}=8$ layers, hidden dimension $D=384$, 6 attention heads, $\sim$15M parameters.
\item \textbf{ViT-Base}: $L_{\mathrm{enc}}=12$ layers, hidden dimension $D=768$, 12 attention heads, $\sim$86M parameters.
\item \textbf{ViT-Large}: $L_{\mathrm{enc}}=24$ layers, hidden dimension $D=1024$, 16 attention heads, $\sim$304M parameters.
\end{itemize}
The model size increase around 4-5x between successive variant. We pretrain each model variant on the same dataset with identical hyperparameters, and the decoder also remains fixed with $L_{\mathrm{dec}}=4$ layers, hidden dimension $D_{\mathrm{dec}}=512$, and 8 attention heads. Figure.\ref{fig:model_capacities} shows cross-domain performance on Widar 3.0 across different model sizes. For both cross-subject and cross-environment evaluations, the results show that the performance has saturated before the model size reaches ViT-Base. Though larger model, for example ViT-Base, has slightly better performance (0.4\% and 0.9\% improvements for cross-subject and cross-environment respectively) than ViT-Small, the gain is marginal compared to the significant increase in computational cost, which is shown in Table.\ref{tab:flops}. Therefore, ViT-Small provides a good trade-off between accuracy and efficiency for Wi-Fi sensing foundation models so far. We can also see the MSE loss decreases as the model size increases, indicating that larger models indeed have stronger learning capabilities. This suggests that the current bottleneck may lie more in data scale and diversity rather than model capacity (as we also discussed in \cref{sec:data_scaling}). Note that the MSE losses for these two experiments are very close to each other; therefore, we only provide the MSE loss for the cross-subject experiment in Figure.\ref{fig:model_capacities} for simplicity.

\subsubsection{Impact of Masking Ratio} \label{sec:mask_ratio}
\begin{figure}[t]
    \centering
    \includegraphics[width=\columnwidth]{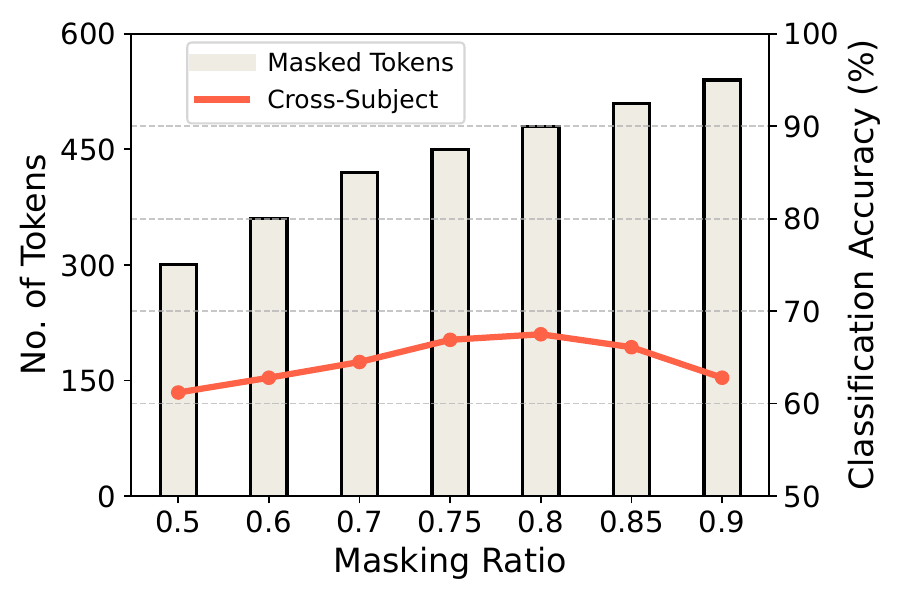}
    \caption{Impact of Masking Ratio}
    \label{fig:masking_ratio}
\end{figure}

We vary the masking ratio from 50\% to 90\% and measure its effect on cross-subject performance. From Figure.\ref{fig:masking_ratio} we can see that higher masking ratios (75--80\%) generally improve cross-domain generalization by forcing the model to infer global structure from limited context, consistent with vision MAE findings. However, extremely high ratios ($\ge$85\%) start to hurt performance because too little spatial temporal information remains for the model to connect with meaningful patterns or global dependencies. Conversely, lower masking ratios ($\le$70\%) lead to reconstruction tasks that are too easy, resulting in less robust representations. We find that 80\% masking provides the best performance while still maintaining efficient pretraining.

\subsubsection{Patch Size and Spatiotemporal Granularity} \label{sec:patch_size}
\begin{figure}[t]
    \centering
    \includegraphics[width=\columnwidth]{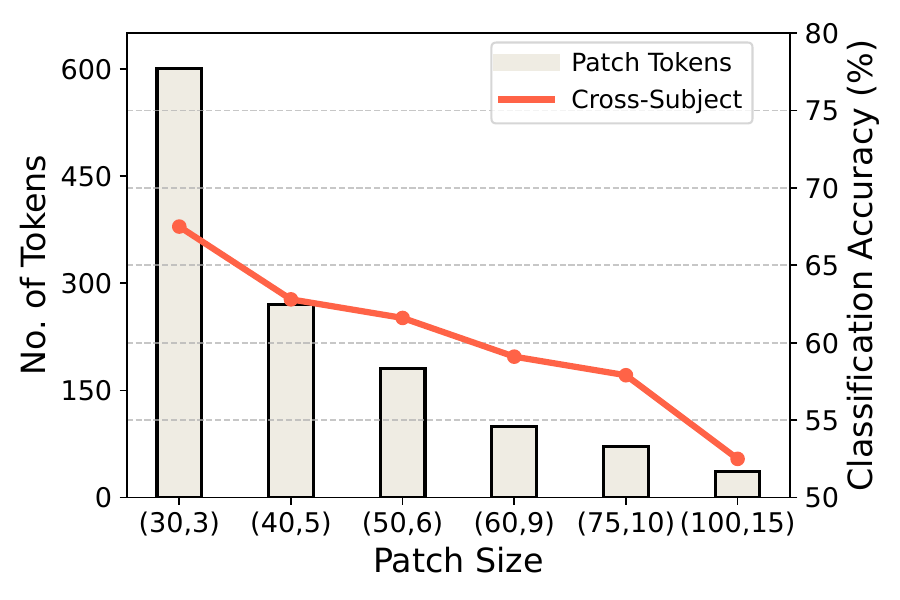}
    \caption{Impact of Patch Size}
    \label{fig:patch_size}
\end{figure}

Patch size determines the granularity of spatiotemporal tokens fed into the transformer encoder, impacting both the model's ability to capture fine-grained motion patterns and computational efficiency. We experiment with patch sizes ranging from (30,3) to (100,15) along the time and frequency dimensions. Figure.\ref{fig:patch_size} shows the cross-subject performance on Widar 3.0 dataset under different patch sizes. It is clear that smaller patch size achieves better performance, as it preserves more detailed temporal dynamics essential for motion recognition. Though smaller patches increase computational costs due to the larger number of tokens (grey bar in the figure), the gains in accuracy justify the trade-off, especially when the patch size is (30,3), which outperforms (40,5) by 4.7\%. 

\subsection{Data Scaling and Diversity}\label{sec:data_scaling}

Similar to how model capacity affects generalization, we study how pretraining data scale and diversity influence cross-domain transfer performance. 

\subsubsection{Impact of Pretraining Data Scale}
Again, we use the Widar 3.0 as the downstream task dataset to study the effects of pretraining data scale. After constructing the pretraining dataset, we pretrain models on varying amounts of CSI samples (from 1\textperthousand \space to 100\%) randomly drawn from the pretraining dataset and evaluate their transfer performance.
Figure.\ref{fig:data_scaling} shows that both cross-subject and cross-environment hand gesture recognition accuracy improves consistently with more pretraining data, exhibiting log-linear scaling trends. Interestingly, our results also show that pretraining is \textbf{not} always beneficial when the data volume is small. For instance, with only 1\textperthousand ($\sim1320$ samples)\space of pretraining data, the pretrained model underperforms the supervised baseline by 5.5\%. When the data scale increases to 1\% ($\sim13200 $ samples), the pretrained model achieves comparable performance with the supervised baseline (blue and red dashed line). This highlights the critical role of large-scale, diverse data in realizing the benefits of foundation models for Wi-Fi sensing. Besides, we cannot see the saturation of performance yet, indicating that even larger datasets could further boost generalization.

\begin{table}[]
    \centering  
    \caption{Cross-domain performance on Widar 3.0 when excluding it from pretraining}
    \resizebox{\columnwidth}{!}{
    \begin{tabular}{@{}cccc@{}}
    \toprule\toprule
    \multicolumn{1}{c|}{Experiment} & Pretraining MSE & Cross-Subject & Cross-Environment \\ \midrule
    \multicolumn{1}{c|}{Supervised} &  \textemdash  &    54.7\%   &    33.5\%      \\ \midrule
    \multicolumn{1}{c|}{Pretrained w/ Widar}   & 0.0148   & 67.5\%  &  41.2\% \\ 
    \rowcolor{black!8}
    \multicolumn{1}{c|}{Pretrained w/o Widar}   & 0.2611   & 65.8\%(8.5\%$\uparrow$)  &  38.6\% \%(5.1\%$\uparrow$)\\ 
    \bottomrule
    \end{tabular}
    }
    \label{tab:exclude_widar}
    \vspace{0mm}
\end{table}

\subsubsection{Role of Data Diversity}
To disentangle scale from diversity, in this evaluation, we construct the pretraining dataset by excluding the entire target downstream dataset (Widar 3.0) to avoid any potential data leakage. This evaluation essentially tests the Out-of-the-Box capabilities of the foundation model by mimicking a more realistic scenario where the foundation model is pretrained on large-scale diverse datasets and then transferred to a completely unseen downstream dataset collected by end users. Table.\ref{tab:exclude_widar} shows the cross-domain performance on Widar 3.0 when excluding it from pretraining. We can see that even without any data from the target downstream dataset during pretraining, the foundation model still significantly outperforms the supervised baseline by 8.5\% and 5.1\% for cross-subject and cross-environment evaluations respectively. Compared to the model pretrained with Widar 3.0 included (Table \ref{tab:hgr_ui_acc}), the performance only drops slightly by 1.7\% and 2.6\% respectively. This indicates that the diversity of pretraining data also plays a critical role along with the scale in enhancing cross-domain generalization for Wi-Fi sensing foundation models. We also show some reconstructed samples in Figure.\ref{fig:generative_vis} (the 4th row). Even though we can see some artifacts, the signal dynamics from the reconstructed CSI can still be clearly visualized.

\subsubsection{Cross-Band and Cross-Device Generalization}

\begin{table}[t]
    \centering
    \caption{Cross-band generalization on WiMANS dataset}
    \label{tab:cross_band}
    \resizebox{\columnwidth}{!}{
        \begin{tabular}{@{} c c c c c c @{}}
            \toprule\toprule
            \multirow{2}{*}{} & \multirow{2}{*}{Method} &\multicolumn{2}{c}{2.4GHz} &\multicolumn{2}{c}{5GHz} \\
            \cmidrule(lr){3-4}\cmidrule(lr){5-6}
            & & {cross-subject} & {cross-env} & {cross-subject} & {cross-env} \\
            \midrule
            & Supervised & 17.8\% & 17.4\% & 57.5\% & 48.2\% \\
            \midrule
            \rowcolor{black!8}
            & Pretrained w/ FT  & 40.2\% (22.4\% $\uparrow$) & 17.7 (0.3\% $\uparrow$) & 62.8\% (5.3\% $\uparrow$) & 56.9\% (8.7\% $\uparrow$) \\
            \bottomrule
        \end{tabular}
    }
    \vspace{5mm}
\end{table}
From Table.\ref{tab:datasets}, we can see that most of Wi-Fi sensing datasets are only collected on 5GHz band, as higher frequency provides better sensitivity to human motions. Even so, 2.4GHz band is still widely used in many existing Wi-Fi sensing works due to its better penetration ability. Therefore, in this section, we investigate how foundation models pretrained on 5GHz data can generalize to 2.4GHz frequency bands. Therefore, we use the WiMANS dataset (2.4 GHz part) as the target downstream dataset and pretrain the foundation model on all the other datasets in 5GHz. We also report the fine-tuning accuracy of WiMANS dataset when only 5GHz data are evaluated. Note that Table.\ref{tab:har_acc} reports the accuracy of WiMANS dataset when both 2.4GHz and 5GHz data are used for pretraining and evaluation. Table.\ref{tab:cross_band} shows that the pretrained foundation model can still improve the performance of both cross-subject and cross-environment human activity recognition on 2.4GHz WiMANS dataset, even though the pretraining data and downstream data are collected on different frequency bands. This indicates that the foundation model has learned some frequency-invariant representations that can generalize across frequency bands.

\begin{table}[]
    \centering  
    \caption{Cross-device generalization on NTU-Fi dataset}
    \resizebox{0.75\columnwidth}{!}{
    \begin{tabular}{@{}ccc@{}}
    \toprule\toprule
    \multicolumn{1}{c|}{Task} & NTU-Fi HAR & NTU-Fi Human-ID \\ \midrule
    \multicolumn{1}{c|}{Supervised} &  92.6\%   &    78.3\%      \\ \midrule
    \rowcolor{black!8}
    \multicolumn{1}{c|}{Pretrained w/ FT} & 95.6\%(3\%$\uparrow$) & 86.1\% \%(7.8\%$\uparrow$)\\ 
    \bottomrule
    \end{tabular}
    }
    \label{tab:cross_device}
    \vspace{0mm}
\end{table}

Additionally, we also investigate how foundation models can generalize across different Wi-Fi devices. From Table.\ref{tab:datasets}, we can see that most of the datasets were collected by Intel 5300 NICs. However, in real-world applications, commercial Wi-Fi APs/routers are more commonly used for Wi-Fi sensing due to their widespread deployment. Therefore, in this evaluation, we use NTU-Fi dataset as the target downstream dataset, which was collected by TP-Link N750. We exclude any datasets collected by this router from the pretraining dataset. Table.\ref{tab:cross_device} shows that the pretrained foundation model can still improve the performance of both human activity recognition and user identification on NTU-Fi dataset even though the pretraining data and downstream data are collected by different Wi-Fi devices. Note that the NTU-Fi dataset comes with predefined train-test splits and lacks explicit domain annotations, so we report accuracy using the provided split without further domain-specific evaluation protocols.


\section{DISCUSSION}
\noindent
In this section, we reflect on the broader implications of our findings and discuss the key factors that influence the scalability and generalization of Wi-Fi foundation models. Our discussion focuses on three main aspects: the need for larger and more diverse Wi-Fi CSI datasets, the development of more informative and transferable Wi-Fi sensing features, and the computational cost associated with large-scale pretraining. Together, these aspects outline the critical challenges and future directions for building scalable, efficient, and robust Wi-Fi foundation models suitable for real-world deployment.
\subsection{Larger-scale Wi-Fi CSI Data.}
In \cref{sec:data_scaling}, we observe a clear trend: exponentially increasing the volume of pretraining data leads to substantial improvements in cross-domain transfer performance. This observation aligns with recent progress in other domains, where data scale has been shown to be the key driver of robust generalization in foundation models. Interestingly, \cref{sec:model_capacity} reveals that performance gains begin to plateau when model capacity increases beyond a certain threshold. This saturation effect indicates that simply scaling up the model size without proportionally increasing the diversity and quantity of training data yields diminishing returns.
The interplay between data volume and model capacity suggests that these two factors must be carefully balanced to achieve optimal performance. Based on the observed scaling trends, we hypothesize that pretraining on datasets several orders of magnitude larger—potentially tens or hundreds of times the current Wi-Fi CSI volume—could yield significantly stronger cross-domain generalization. However, achieving such scale faces practical obstacles. Existing datasets remain fragmented, inconsistent, and difficult to aggregate due to variations in device configurations, environments, and collection protocols. Additionally, large-scale in-the-wild Wi-Fi CSI collection is hindered by privacy concerns and logistical challenges. Future research could explore synthetic data generation, domain simulation, or federated learning frameworks to leverage distributed data sources while preserving user privacy and ensuring data diversity.
\subsection{Better Wi-Fi Sensing Features.}
The success of foundation models in vision, language, and multimodal domains is largely attributed to their ability to learn rich and generalizable representations that capture the underlying structure of the data. In the context of Wi-Fi sensing, however, existing CSI representations may not fully exploit the rich physical and temporal characteristics of wireless propagation. For instance, in our current implementation, phase information is discarded because certain datasets provide only processed amplitude values. This constraint likely limits the model’s capacity to learn fine-grained motion and spatial features that are critical for robust transfer across domains.
Future work should investigate more comprehensive and expressive CSI representations that better capture both amplitude and phase dynamics, as well as their spatial-temporal correlations. Additionally, leveraging physics-informed priors or hybrid feature encoders that integrate domain knowledge with data-driven representations may further enhance model interpretability and robustness. By systematically improving the quality and expressiveness of input features, Wi-Fi foundation models could learn more transferable representations capable of generalizing across a wide range of environments and hardware configurations.
\begin{table}[] \centering \caption{Computational cost (FLOPs) of ViT variant during pretraining and evaluation stages.} \resizebox{\columnwidth}{!}{ \begin{tabular}{@{}ccccc@{}} \toprule\toprule \multicolumn{1}{c|}{Method} & ViT-Tiny & ViT-Small & ViT-Base & ViT-Large \\ \midrule \multicolumn{1}{c|}{Pretraining} & 9.36GFLOPS & 12.19GFLOPS & 29.45GFLOPS & 81.73GFLOPS \\ \midrule \multicolumn{1}{c|}{Evaluation} & 3.26GFLOPS & 17.19GFLOPS & 102.58GFLOPS & 364.18GFLOPS \\ \bottomrule \end{tabular} } \label{tab:flops} \vspace{0mm} \end{table}

\subsection{Computational Cost and Efficiency.}
While our experiments demonstrate the feasibility of training foundation models on Wi-Fi CSI data, the computational cost associated with large-scale pretraining remains a major concern. As shown in Table~\ref{tab:flops}, the FLOPs of ViT variants grow rapidly with model size, reflecting substantial increases in both training and inference costs. As model capacity and data volume continue to scale, so too do training time, energy consumption, and hardware requirements.
To address these challenges, future work should focus on improving computational efficiency through lightweight architectures, knowledge distillation, and parameter-efficient fine-tuning strategies. Model compression techniques, such as pruning or quantization, may further reduce inference overhead, enabling deployment on edge devices with limited resources. Additionally, exploring adaptive training pipelines—such as progressive pretraining or curriculum-based data sampling—could provide an effective balance between performance and efficiency. Together, these directions can make large-scale Wi-Fi foundation models more practical and sustainable for widespread use.
\section{RELATED WORK}
Recent years have witnessed remarkable progress at the intersection of wireless communications, sensing, and machine learning. In this section, we review three key areas relevant to our work:
Wireless Foundation Models, Wi-Fi and Wireless Sensing, and recent developments in self-supervised learning.

\subsection{Wireless Foundation Models.} Inspired by major advances in language and vision \cite{devlin2019bert, brown2020language, dosovitskiy2020image, radford2021learning,he2022masked}, foundation models are reshaping wireless research by replacing traditional task-specific networks with general-purpose encoders pretrained on large-scale radio data, which can be reused across various downstream communication and sensing tasks. LWM\cite{alikhani2024large} first introduces the concept of wireless foundation models by pretraining on a large synthetic MIMO dataset \cite{alkhateeb2019deepmimo} and demonstrates outperforming results on communications tasks such as beam selection and LOS/NLOS classification. Following LWM, several works have further explored the potential of wireless foundation models on wireless communications tasks such as channel estimation\cite{yang2025wirelessgpt}, spectrum management\cite{zhou2025spectrumfm} and signal understanding\cite{luo2025emind}. However, due to the lack of real radio-sensing datasets, these works primarily focus on communication tasks, leaving the potential of wireless foundation models in wireless sensing largely unexplored. To overcome the dataset dilemma,  FM-Fi\cite{weng2024large} and its successor\cite{weng2025fm} utilize contrastive learning (CLIP) to align multimodal data such as vision and text in conjunction with the wireless siganls to pretrain wireless foundation model/RF encoder for sensing tasks. There are also works that pretrain a versatile, cross-modal foundation model to further improve sensing performance \cite{chen2024x}.

\subsection{Wi-Fi and Wireless Sensing.}
Recent decades have witnessed significant advancements in wireless sensing technologies, leveraging the ubiquitous presence of radio signals, such as Wi-Fi, for various sensing applications \cite{ma2018signfi,shi2017smart,islam2024wi,zhang2020gaitid,yang2022rethinking,huang2024wimans}. Wi-Fi sensing research nowadays mainly focuses on improving the generalization of Wi-Fi sensing systems under diverse real-world conditions. For instance, Wi-Fi-based human activity/hand gesture recognition have been extensively studied, with methods evolving from traditional machine learning techniques based on the RSSI \cite{abdelnasser2015wigest} to deep learning approaches with finer-grained and more robust representations, e.g. BVP \cite{zheng2019zero} and DFS \cite{xiao2021onefi}. Wi-Fi sensing has also been applied to emerging applications, such as vital sign monitoring \cite{liu2015tracking, hu2024m,li2024spacebeat}, hand/body pose estimation \cite{ren2022gopose,ji2023construct,jiang2020towards, yan2024person}, and even emotion recognition \cite{zhao2016emotion}. Beside Wi-Fi, other wireless technologies such as RFID, UWB, and mmWave have also been explored for these sensing applications. These technologies offer different trade-offs in terms of range, accuracy, and cost, making them suitable for different scenarios. However, all of wirelss sensing tech suffer from similar challenges, which attract research interests in developing robust and generalizable sensing algorithms.

\subsection{Self-Supervised Learning.}
Self-supervised Learning (SSL) founds the roots of foundation models by allowing models to explore useful representations from large-scale unlabeled data. SSL methods can be generally categorized into contrastive learning and masked modeling. Both approaches treat the input data as supervision signals by designing pretext tasks to guide the models to learn from inherent features within the data itself. Contrastive learning methods, such as SimCLR \cite{chen2020simple} and MoCo \cite{he2020momentum}, construct positive and negative samples tactfully to learn representations by calculating the contrastive loss between positive pairs and negative pairs, effectively capturing discriminative features. Later works such as BYOL \cite{grill2020bootstrap}, SimSiam \cite{chen2021exploring} and DINO \cite{chen2021exploring} completely ditch the need of negative samples and rely solely on positive pairs to learn meaningful representations. In terms of masked modeling methods, like BERT \cite{devlin2019bert} in natural language processing and MAE \cite{he2022masked} in computer vision, apply masking strategies to enforce models to reconstruct masked parts of the input data, which let the model learn contextual information. While the random masking used in these methods has achieved decent results, recent efforts on finding better masking strategies, such as reinforcement learning based adaptive masking \cite{bandara2023adamae} and semantic guided masking \cite{li2022semmae}, have shown further performance improvements.

\subsection{Summary.}
In summary, prior research in wireless foundation models has laid the groundwork for unifying diverse wireless communication and sensing tasks through large-scale pretraining, yet real-world generalization—especially in sensing—remains limited by data scarcity and modality gaps. Advances in Wi-Fi and wireless sensing demonstrate the promise of radio-based perception but continue to face challenges of robustness and domain adaptation. Meanwhile, progress in self-supervised learning offers powerful strategies for exploiting large unlabeled datasets, providing a natural pathway toward scalable and generalizable wireless foundation models. Building upon these insights, our work bridges these areas by leveraging self-supervised pretraining on large-scale, heterogeneous Wi-Fi data to advance cross-domain sensing performance.

\section{CONCLUSION}


In this work, we presented a large-scale study on the potential of foundation model pretraining for Wi-Fi sensing. By aggregating and harmonizing 14 heterogeneous Wi-Fi CSI datasets across multiple frequency bands, devices, and environments, we demonstrated that large and diverse data pretraining can substantially enhance cross-domain generalization. Our results reveal clear scaling laws: unseen domain performance improves log-linearly with data scale, while increasing model capacity yields only marginal benefits under current data limitations. These findings highlight that data diversity, rather than model complexity, is the primary bottleneck for achieving robust and generalizable Wi-Fi sensing. Beyond the quantitative gains observed across multiple downstream tasks—including activity recognition, gesture classification, and user identification—our analysis provides a foundation for future research toward unified, deployable wireless sensing models. We envision that continued efforts in large-scale data collection, multimodal integration, overcoming data dilemma, potentially through synthetic data or federated learning, and design better features representations, will ultimately enable reliable, camera-free sensing systems ready for real-world use.

\balance
\bibliographystyle{ACM-Reference-Format}
\bibliography{reference}

\end{document}